\documentclass{article}

\usepackage[nonatbib, preprint]{neurips_2023}

\usepackage[utf8]{inputenc} %
\usepackage[T1]{fontenc}    %
\usepackage{url}            %
\usepackage{booktabs}       %
\usepackage{amsfonts}       %
\usepackage{nicefrac}       %
\usepackage{microtype}      %
\usepackage{xcolor}         %
\usepackage[pdftex]{graphicx}
\usepackage{gensymb}
\usepackage{enumerate}
\usepackage{paralist}
\usepackage{amsmath}
\usepackage{multirow}
\usepackage{bbm}
\usepackage{wrapfig}
\usepackage{graphicx}
\usepackage{subcaption}
\usepackage{tikz}
\usepackage{caption}
\usepackage{adjustbox}

\newcommand{\bluediamond}{%
  \begin{adjustbox}{height=1.5ex,valign=center}
    \begin{tikzpicture}[baseline=-0.5ex, scale=1.5]
    \fill[blue] (0,0) -- (0.5ex,1ex) -- (1ex,0) -- (0.5ex,-1ex) -- cycle;
    \end{tikzpicture}
  \end{adjustbox}%
}

\newcommand{\reddiamond}{%
  \begin{adjustbox}{height=1.5ex,valign=center}
    \begin{tikzpicture}[baseline=-0.5ex, scale=1.5]
    \fill[red] (0,0) -- (0.5ex,1ex) -- (1ex,0) -- (0.5ex,-1ex) -- cycle;
    \end{tikzpicture}
  \end{adjustbox}%
}

\DeclareRobustCommand{\bluearrowright}{%
  \begin{tikzpicture}[baseline=-0.5ex, scale=1.5]
    \path[draw=blue,-latex,line width=0.8pt] (-0.8ex,0) -- (0.8ex,0);
    \path[draw=blue,latex-,line width=0.8pt] (0.8ex,0) -- (-0.8ex,0);
  \end{tikzpicture}%
}

\DeclareRobustCommand{\bluearrowleft}{%
  \begin{tikzpicture}[baseline=-0.5ex, scale=1.5]
    \path[draw=blue,-latex,line width=0.8pt] (0.8ex,0) -- (-0.8ex,0);
    \path[draw=blue,latex-,line width=0.8pt] (-0.8ex,0) -- (0.8ex,0);
  \end{tikzpicture}%
}

\usepackage[pagebackref=true,breaklinks=true,colorlinks,bookmarks=false]{hyperref}

\definecolor{almond}{rgb}{0.94, 0.6, 0.4}
\def\bmatrix#1{\left[ \begin{matrix} #1 \end{matrix} \right]}  %

\title{POPE: 6-DoF Promptable Pose Estimation of Any Object, in Any Scene, with One Reference}

\author{
  \textbf{Zhiwen Fan\textsuperscript{1}\thanks{Equal contribution}\,, Panwang Pan\textsuperscript{2$\ast$}, Peihao Wang\textsuperscript{1}, Yifan Jiang\textsuperscript{1}},  \\
  \textbf{Dejia Xu\textsuperscript{1}, Hanwen Jiang\textsuperscript{1}, Zhangyang Wang\textsuperscript{1}}\\
  {\textsuperscript{1}University of Texas at Austin}, {\textsuperscript{2}Bytedance}
  \\ \small{\texttt{\{zhiwenfan,atlaswang\}@utexas.edu}}
}

\begin{document}

\maketitle

\begin{figure}[!ht]
  \centering
  \vspace{-6mm}
  \includegraphics[width=1\linewidth]{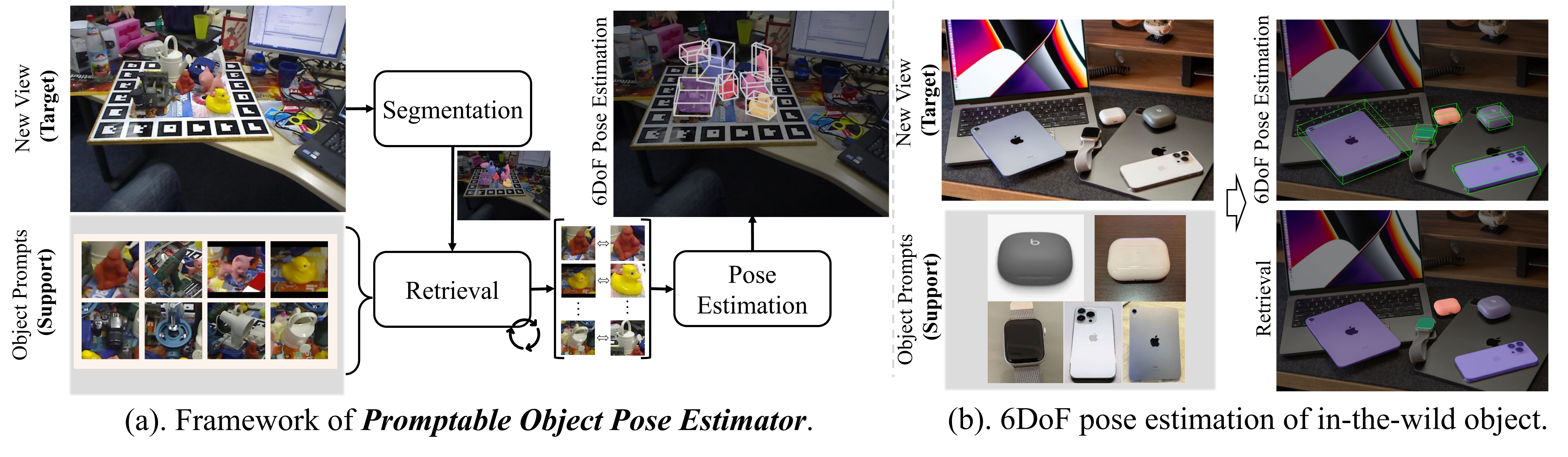}
  \vspace{-6mm}
  \caption{Promptable object Pose Estimator (\textbf{POPE}) is a zero-shot object 6DoF pose estimation method, which predicts the relative pose between given object prompts (\textbf{support view}) and the object in any new view (\textbf{target view}). Our framework recognizes the object prompts in the target under any scene (as shown in a cluttered scene in (\textbf{a})) and estimates the relative pose for any category with only one support image. POPE also exhibits capability on any object pose estimation (as shown in (\textbf{b})).
  \vspace{-3mm}
  }
  \label{fig:teaser}
\end{figure}

\begin{abstract}
Despite the significant progress in six degrees-of-freedom (6DoF) object pose estimation, existing methods have limited applicability in real-world scenarios involving embodied agents and downstream 3D vision tasks. These limitations mainly come from the necessity of 3D models, closed-category detection, and a large number of densely annotated support views. 
To mitigate this issue, we propose a general paradigm for object pose estimation, called \textbf{\textit{\underline{P}romptable \underline{O}bject \underline{P}ose \underline{E}stimation (POPE)}}. The proposed approach POPE enables zero-shot 6DoF object pose estimation for \underline{any target object in any scene}, while only a single reference is adopted as the support view.
To achieve this, POPE leverages the power of the pre-trained large-scale 2D foundation model, employs a framework with hierarchical feature representation and 3D geometry principles. Moreover, it estimates the relative camera pose between object prompts and the target object in new views, enabling both two-view and multi-view 6DoF pose estimation tasks. Comprehensive experimental results demonstrate that POPE exhibits unrivaled robust performance in zero-shot settings, by achieving a significant reduction in the averaged Median Pose Error by \textbf{52.38\%} and \textbf{50.47\%} on the LINEMOD~\cite{hinterstoisser2013model} and OnePose~\cite{sun2022onepose} datasets, respectively. We also conduct more challenging testings in causally captured images (see Figure~\ref{fig:teaser}), which further demonstrates the robustness of POPE. Project page can be found with \url{https://paulpanwang.github.io/POPE/}.

\end{abstract}

\begin{figure}[t]
  \centering
  \includegraphics[width=1\linewidth]{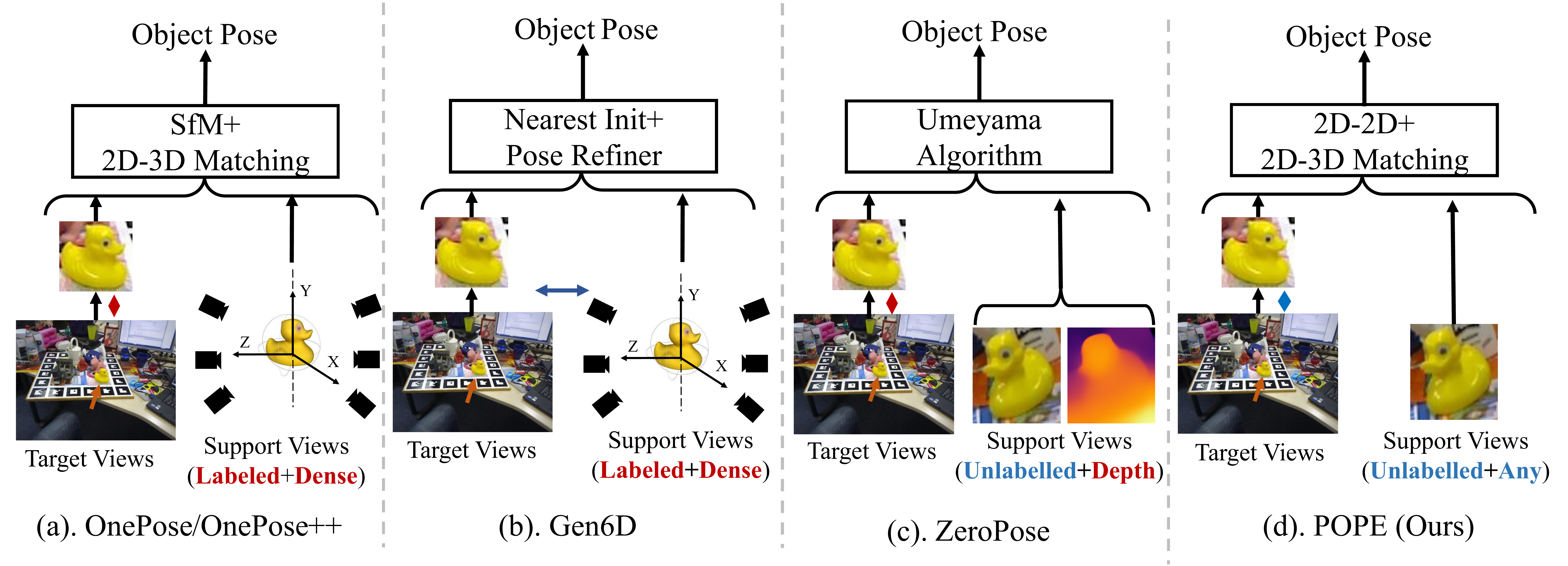}
  \vspace{-0.25in}
  \caption{\textbf{Comparing POPE with previous frameworks.} We provide a detailed comparison between POPE and prior works, including (a) OnePose/OnePose++~\cite{sun2022onepose,he2023onepose++} which relies on a large number of posed support views and the corresponding bounding box; (b) Gen6D~\cite{gen6d} that replace 2D-3D matching pipeline with a refiner network; and (c) ZeroPose~\cite{goodwin2022zero} which further utilized depth maps. Different from all these methods, the proposed method POPE eliminates the need for densely annotated support views and enables accurate object retrieval in new viewpoints without relying on any assumptions about the object's category. 
  Here, \reddiamond{} denotes a close-category detector, \bluearrowleft{}\bluearrowright{} means a correlation-based detector, and \bluediamond{} denotes a open-world detector.
  \vspace{-4mm}}
  \label{fig:method_comparison}
\end{figure}

\section{Introduction}
Robotic systems and augmented reality/virtual reality (AR/VR) applications have become ubiquitous across numerous industries, facilitating the execution of intricate tasks ot offering immersive user experiences. Describing the status of objects, particularly their six degrees-of-freedom (6DoF) poses, is a crucial step towards achieving in-depth scene understanding and delicate interactions. More importantly, given the diverse nature of real-world scenarios, it is essential to have a method that can operate on arbitrary object assets.

However, enabling object 6DoF pose estimation on unseen objects using simple and easy-to-obtained references is challenging.
Traditional instance-level~\cite{kehl2017ssd,tekin2018real,xiang2017posecnn,zakharov2019dpod,li2019cdpn,peng2019pvnet,labbe2020cosypose} or category-level~\cite{wang2019normalized,ahmadyan2021objectron,chen2020category} pose estimators exhibit limitations in handling diverse objects, as they are specifically designed for particular instances or categories.
These design principles restrict their generalization capabilities to unseen instances or categories during testing, due to their reliance on CAD models or a well-defined category-level canonical space. Later, tremendous efforts have been devoted to addressing the aforementioned challenges by adopting structure-from-motion (SfM~\cite{schonberger2016structure}) techniques~\cite{sun2022onepose,he2023onepose++}, reducing the number of support views~\cite{gen6d}, or leveraging depth maps and self-supervised trained Vision Transformers~\cite{goodwin2022zero}. A detailed visual comparison is summarized in Figure~\ref{fig:method_comparison}.

A straightforward way to accomplish 6DoF object pose estimation with a single support view is to estimate relative poses~\cite{sinha2022sparsepose, zhang2022relpose} by performing 2D-2D matching between query and reference images. However, dense matching on arbitrary objects is highly unstable, especially for wide-baseline camera views or a clustered background. Besides the difficulties in image matching, another substantial issue in real-world scenes arises from the potential for heavy occlusion of the target object, which makes it hard to be detected. Previous methods propose to adopt off-the-shelf detectors~\cite{jocher2020yolov5} for specific instances/categories, or design a correlation-based object detector on a small scale dataset~\cite{gen6d}. Consequently, their robustness when dealing with novel objects in diverse scenes are not guaranteed.

To tackle the obstacles of open-world detection on arbitrary target objects and the robust 2D-2D matching, a promising avenue is leveraging the power of the foundation model that is trained on a vastly large-scale dataset. Recently, the community has witnessed the emerging properties of these foundation models on few-shot or even zero-shot generalization, crossing from language~\cite{devlin2018bert,brown2020language} to vision~\cite{oquab2023dinov2,caron2021emerging,kirillov2023segment}.

These advancements have shed light on the under-explored problem of zero-shot object pose estimation - the tantalizing possibility of making no assumption on the object category and using only one reference image. Specifically, the newly arising capability of performing zero-shot segmentation across various image data domains~\cite{kirillov2023segment} and non-parametric instance-level recognition~\cite{oquab2023dinov2} have shown potential in addressing these challenges.

In this paper, we introduce a novel task named \textbf{\textit{Promptable Object Pose Estimation}} to tackle the challenge of estimating the 6DoF object pose between the given object prompts (a single image for each instance, used as support) and any new captured viewpoint with complex backgrounds (target). Our proposed model, called \textbf{\textit{POPE}}, consists of four main features in one unified pipeline: (i) \textit{Segment Objects} generates a set of valid segmentation proposals for any image at a new viewpoint; (ii) \textit{Retrieve Objects} constructs object-level matching between object prompts and segmented object proposals at two views; (iii) \textit{Pose Objects} estimate the relative pose by utilizing the matched correspondences between paired object images; and (iv) \textit{Online Refinement for Arbitrary View-number} triggers a coarse-to-fine pose estimation process with efficient 2D-2D global matching and 2D-3D local matching for retrieved objects, on newly target views. We outline our contributions below:

\begin{itemize}
\item We establish a new and challenging task: \textbf{\textit{Promptable Object Pose Estimation}}, which aims to estimate the pose of an object in wild scenarios, with no assumptions on object category and using only one reference image.

\item To tackle this problem, we propose a 6DoF pose foundation model, \textbf{POPE}, that seamlessly integrates the power of pre-trained foundation models and 3D geometry principles for high-quality segmentation, hierarchical object retrieval, and robust image matching, to enable accurate object pose estimation in diverse and uncontrolled environments.

\item For evaluation, we introduce a large-scale test dataset containing diverse data sources. POPE outperforms existing generalizable pose estimators and demonstrates remarkable effectiveness for both promptable pose estimation and downstream 3D-vision tasks.
\end{itemize}

\section{Related Works}

\paragraph{Large-scale Pre-trained 2D Foundation Models.}
Models trained on large-scale datasets, demonstrating the scaling effect with data-parameter balance, are regarded as foundation models. Recently, we witnessed that the foundation models~\cite{brown2020language} demonstrated strong generalization capability, serving as the base model in a wide range of tasks~\cite{brown2020language}. For example, CLIP~\cite{radford2021learning} utilizes contrastive learning to construct a joint embedding space of text and image  modalities.
Similarly, self-supervised models such as DINO~\cite{caron2021emerging} and 
DINOv2~\cite{oquab2023dinov2} show emerging properties for learning robust visual features.
Segment-Anything Model (SAM)~\cite{kirillov2023segment} demonstrates promptable segmentation ability that supports interactive segmentation with visual promptings such as points and bounding boxes.
In this paper, we achieve the goal of promptable object pose estimation by harnessing the power of foundation models. We build a system that integrates the essence of SAM and DINO to help POPE handle cluttered real scenes by performing dense segmentation and instance-level matching.
\vspace{-4mm}
\paragraph{Generalizable Object Pose Estimator.}
Early approaches for estimating the 6DoF pose of objects build instance-level~\cite{yen2021inerf,hodan2020epos,di2021so} or category-level~\cite{wang2019normalized,chen2020category,wen2021disentangled,deng2022icaps,lin2021sparse,chen2020category,chen2021fs,lin2021dualposenet,tian2020shape,di2022gpv,goodwin2022zero} frameworks. They usually require perfect instance-specific CAD models or well-established canonical space of specific categories. Thus, the methods only work on specific instances and categories. They cannot generalize to novel instances/categories that are unseen during training.
The recent advances in generalizable object pose estimators can be divided into two categories based on whether a 3D model is utilized. One line of work adopts high-quality 3D objects through shape embedding~\cite{xiao2019pose,pitteri20203d,dani20213dposelite}, template matching~\cite{hinterstoisser2011multimodal,balntas2017pose,wohlhart2015learning,sundermeyer2020multi} and rendering-and-comparison approaches~\cite{li2018deepim, zakharov2019dpod, okorn2021zephyr, busam2020like}.
The other approaches instead aim to avoid the need of 3D objects and utilize depth map~\cite{park2020latentfusion}, object mask~\cite{yen2020inerf,park2020latentfusion,lin2022parallel} and reference images~\cite{gen6d,sun2022onepose,he2023onepose++}.
Specifically, Gen6D~\cite{gen6d} first detects the target object and  initializes a pose estimate from dense reference views. Then, Gen6D refines the pose using feature volume and a 3D neural network.
OnePose ~\cite{sun2022onepose} and OnePose++~\cite{he2023onepose++} construct a sparse point cloud from the RGB sequences of all support viewpoints and then determine the object poses by matching the target view with the sparse point cloud. However, these works still require dense support views, i.e. $\geq$32 views, where each view needs to be annotated with ground-truth poses. We argue the requirement of dense support views is not practical for real-world applications. To this end, we propose the paradigm of promptable pose estimation, where we only use one support view as the reference. We turn the 6DoF object pose estimation task into relative pose estimation between the retrieved object in the target view and the support view. Thus, we do not have any hypothesis of object category, achieving generalizable object pose estimation.
\vspace{-4mm}
\paragraph{Two-view Object Pose Estimation.}

The methods of estimating the relative camera pose between two views can be classified into two categories: i) correspondence-based methods, and ii) direct pose regression methods. The correspondence-based methods establish cross-view pixel-level correspondences, and the pose can be recovered by solving the fundamental matrix~\cite{Mishchuk2017WorkingHT}. The methods establish the correspondences based on hand-crafted features, e.g. SIFT~\cite{lowe2004distinctive}, and SURF~\cite{bay2006surf}, or using learned features~\cite{sarlin2020superglue, sun2021loftr, Jiang2021COTRCT, Li2022PracticalSM, Lindenberger2021PixelPerfectSW, Efe2021DFMAP}. Some of the methods also incorporate robust estimation methods~\cite{Beardsley19963DMA},  or the synergy between shape reconstruction and pose estimation~\cite{Jin2020ImageMA}. 
Another category of methods learns cues for pose estimation in an end-to-end manner~\cite{zhang2022relpose, cai2021extreme, rockwell20228, Jiang2022FewViewOR}. For example, RelPose~\cite{zhang2022relpose} builds an energy-based framework for handling pose ambiguity. The 8-Point Transformer~\cite{rockwell20228} incorporates the inductive bias of the 8-point algorithm into transformer designs. FORGE~\cite{Jiang2022FewViewOR} leverages 3D feature volumes to alleviate the ambiguity of learning on 2D features. In this work, we stick to the classic correspondence-based method because of its better generalization ability on novel instances/categories. Different from prior works establishing image-level correspondence (matching the support image with the entire target image)~\cite{sarlin2020superglue, sun2021loftr}, we propose a coarse-to-fine paradigm. We first build instance-level correspondence by matching the prompt object (shown in the support image) with segmented object instances in the target image, which identifies the highly possible regions of the prompt object. Then we establish fine-grained dense correspondence between the support image and the identified regions in the target image, which avoids noisy matching with cluttered background regions.

\section{Propmtable Object Pose Estimation Task}
Generalizable 6DoF object pose estimators play a crucial role in robotics and 3D vision tasks by accurately determining the position and orientation of novel objects in 3D space, without the need for fine-tuning. 

However, current methods~\cite{gen6d,sun2022onepose,he2023onepose++,goodwin2022zero,yen2020inerf} have limitations.
They can only handle cases where an off-the-shelf detector is used for closed-category object separation from the background~\cite{sun2022onepose,he2023onepose++,goodwin2022zero}.
Additionally, the number of support views required for a robotics system to grasp an object is often uncertain due to occlusions, object appearance variations, and sensor limitations~\cite{yen2020inerf}. 
Furthermore, the tedious requirement of pose annotation~\cite{gen6d,sun2022onepose,he2023onepose++} or depth maps~\cite{goodwin2022zero} in the support view makes it challenging to  scale up and generalize to various scenes.
These limitations hinder the deployment of existing pose estimators in diverse and uncontrolled scenes. 
To address these challenges, we propose to decompose the 6DoF object pose estimation problem into relative object pose estimation. This approach reduces the reliance on absolute pose annotation and allows for easy extension from two-view to multiple-view scenarios. Moreover, we introduce an \textit{Open-world Detector} that is category-agnostic and robust to occlusion and pose variation.

\subsection{Task Definition}
We introduce a novel task of \textbf{\textit{Propmtable Object Pose Estimation (POPE)}}.
The primary goal of this task is to estimate the relative poses for all objects in one scene image according to a series of (single-view) reference image prompts.
Specifically, our POPE model receives an arbitrary scene image and a sequence of arbitrary reference images as the input.
As the output, POPE simultaneously detects all the objects from the scene and annotates their poses according to the references.

\paragraph{Why Promptable?} The use of object prompts allows for higher interactivity and flexibility, enabling end users to indicate their interest in specific objects through prompts such as object images or even abstract sketches. The promptable setting eliminates the reliance on predefined categories or assumptions regarding the size and shape of objects, resulting in a more generalizable approach that can be applied to any object as long as it is included in the set of object prompts.
\paragraph{Why Single-View Prompt?}
We argue that in most user cases, only single-image references are presented and prefered.
On the one hand, consistent images captured for the same object from different angles barely exist in the wild and web collection.
On the other hand, estimating 6DoF pose with multiple views requires additional calibration of the reference views resulting in a chicken-egg problem.
Enabling high-performance two-view geometry also frees the robotic agent from acquiring a CAD model and benefits 3D reconstruction with fewer views.
Despite estimating the poses through only one reference view being a challenging setting, fortunately, it can be endowed with the prevalent foundation models which enable robust feature representation for both detection and matching. 
In addition, single-reference pose estimation can be served as a starting point for multi-view geometry.
Our POPE pipeline can be seamlessly integrated into a multi-view progressive reconstruction pipeline, which consistently boosts pose estimation and reconstruction accuracy starting with a set of unposed images from scratch.

\subsection{Preliminary of Two-view Pose Estimation.}
The task of estimating the relative camera poses 
from two separate images, without a 3D CAD model, is referred to as two-view object pose estimation. Classic geometric vision theory suggests that the camera poses and depth maps can be computed from image matching points alone, without any additional information~\cite{longuet1981computer}.

Given a set of image matching points $\mathbf{x}_i$ and $\mathbf{x}^\prime_i$ in homogeneous coordinates, along with a known camera intrinsic matrix $\mathbf{K}$, the task of two-view object pose estimation is to find the camera rotation matrix $\mathbf{R}$, translation vector $\mathbf{t}$, and corresponding 3D homogeneous point $\mathbf{X}_i$. The goal is to satisfy the equations $\mathbf{x}_i = \mathbf{K}\bmatrix{\mathbf{I}|\mathbf{0}}\mathbf{X}_i$ and $\mathbf{x}_i^\prime = \mathbf{K}\bmatrix{\mathbf{R}|\mathbf{t}}\mathbf{X}_i$ for all $i$. A classical method to solve this problem consists of three steps: computing the essential matrix $\mathbf{E}$ from the image matching points, extracting the relative camera pose $\mathbf{R}$ and $\mathbf{t}$ from $\mathbf{E}$, and triangulating the matching points to get $\mathbf{X}_i$. The essential matrix can be solved using at least 5 matching points~\cite{nister2004efficient}, and $\mathbf{R}$ and $\mathbf{t}$ can be computed from $\mathbf{E}$ using matrix decomposition. There is a scale ambiguity for relative camera pose estimation, and the 3D point $\mathbf{X}_i$ can be computed with a global scale ambiguity.

\subsection{Modular Approach to Zero-shot Promptable Object Pose Estimation}

Directly applying a two-view image matching framework between a prompt image and a complex target containing the same object is prone to failure. This is because a complex scene can have numerous noisy matches, especially when limited to only two observations. Hence, in this paper, we propose a modular approach to address this problem by breaking it down into multiple steps. First, we formulate an \textit{Open-world Detector} that segments and identifies the queried object prompts in the target image. Next, we establish correspondences with new views, refining incorrect object retrievals and solving the task of relative pose estimation.

\vspace{-1mm}
\textbf{Open-world Object Detector.} 
In this paper, we propose a robust and general detector that conditions on the user-provided object prompt image $I_P$ and the image in the target view $I_T$, without making any assumptions about object categories.
The proposed detector aims to obtain the matched object mask in the target view, by generating all $K$ valid masks $\mathcal{M}=\{m^1,m^2,...,m^K\}$ within $I_T$ using automatic object mask generation from a segmentation model\cite{kirillov2023segment}, and retrieving the masked object image with the best global image properties.
Specifically, we generate densely uniform points on the image lattice as prompts for promptable segmentation model (SAM)~\cite{kirillov2023segment} to obtain $\mathcal{M}$, which represents the object segments.
The next goal is to retrieve the masked object image in the target view $I_T$ by establishing the relationship between the object prompt image $I_P$ and the masked object image set $\mathcal{I}_T^K=\{I_T^1, I_T^2,..., I_T^K\}$, given one object prompt image $I_P$ and $K$ object segments in the target. 
However, we cannot guarantee that the image pairs have enough texture~\cite{chen2022deep} or sufficient image content overlapping for local feature matching of the open-world objects.
Inspired by recent progress in self-supervised pre-trained Vision Transformer (ViT) models~\cite{caron2021emerging}, we employ the retrieval augmented data engine in the DINO-v2 model~\cite{oquab2023dinov2} to perform robust global retrieval. 
Here, we utilize the embedded [CLS] token to capture global image properties and construct a cosine similarity matrix of shape $1 \times K$ via the inner product of the [CLS] tokens: $\mathcal{S}\left(P, T, k\right) = \langle{CLS}_{P}, {CLS}_{T}(k)\rangle$, which reveals the object relationship between the prompt image $I_P$ and the $k_{th}$ masked image in set $\mathcal{I}_T^K$.
By finding the highest score within the matrix, we retrieve the matched image of the same object in two views. 
Moreover, extending from a single prompt image to multiple ones (e.g., $M$) can be easily achieved by scaling up the similarity matrix to $M \times K$.

\begin{wrapfigure}{r}{0.45\linewidth}
    \centering
    \includegraphics[width=1\linewidth]{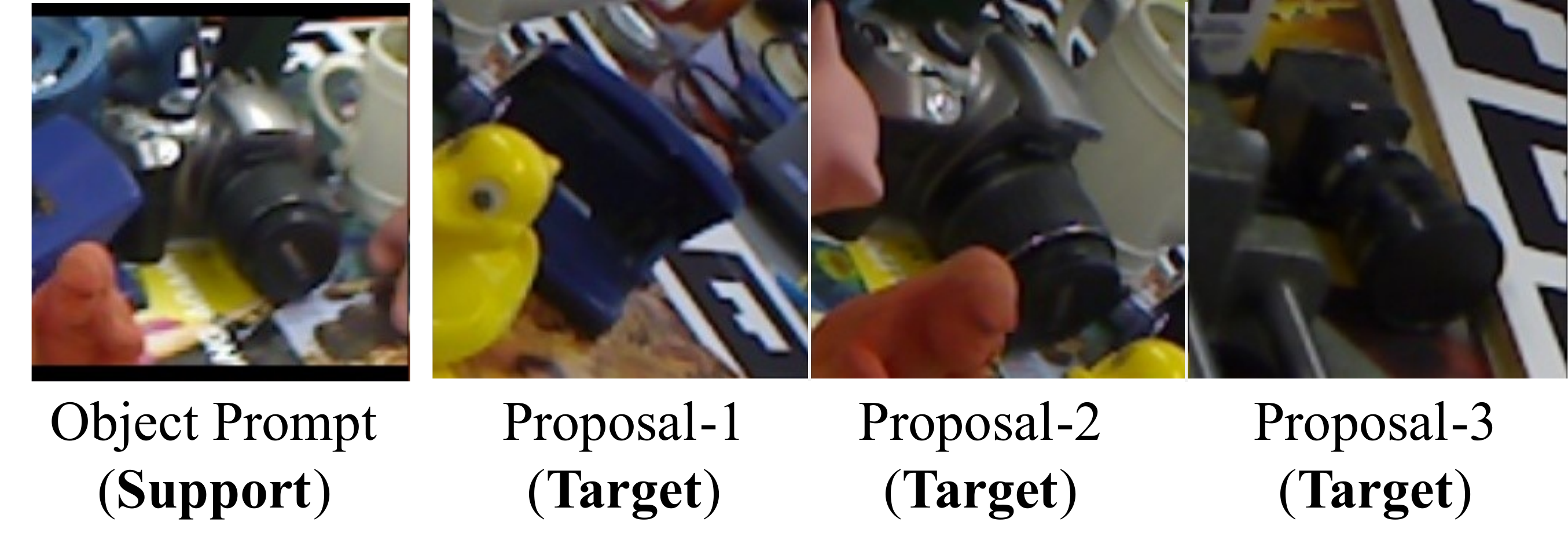}
    \vspace{-0.25in}
    \caption{\textbf{Failed matches.} Relying solely on the similarity score of the [CLS] token for global representation can lead to inaccurate matches, especially in clustered scenarios. This motivates us to incorporate local descriptor information for improved retrieval.}\label{fig:failed_match}
    \vspace{-6mm}
\end{wrapfigure}

\vspace{-4mm}
\paragraph{Hierarchical Retrieval Refinement with Local Descriptors.}
However, despite being trained on a large-scale dataset, DINO-v2 may generate high similarity scores for objects with similar appearances, resulting in erroneous global object-level retrieval (last column of Figure~\ref{fig:failed_match}). 
This, in turn, can negatively impact the accuracy of the pose estimation stage. 
In order to address this issue, we propose a fine-grained approach that incorporates local descriptors to enhance the retrieval process and provide more reliable object matches.
Specifically, we leverage local descriptors to summarize the similarities of local visual patterns, including edges, corners, and textures. These descriptors complement the potentially erroneous retrievals obtained solely from global representations.
To implement this approach, we consider the Top-K proposals generated by DINO-v2, 
ranking the similarity scores in descending order. We then establish image correspondences using a transformer-based local feature estimation framework~\cite{sun2021loftr} 
when \underline{using natural RGB images as prompt}. The predicted confidence matrix $\mathcal{P}_c$ represents the matching probabilities for all correspondences.
To determine the confidence level of the matches, we introduce a confidence criterion based on a threshold value $\sigma$. We select and count the matches with confidence scores higher than the threshold in the total number of matches $n$. This criterion is defined as $\text{Criteria} = \frac{1}{n} \sum_{i=1}^{n} \mathbbm{1}(c_i \geq \sigma)$, where $c_i$ denotes the confidence score of the $i$-th match, and $\mathbbm{1}$ is the indicator function that returns $1$ if its argument is true, and $0$ otherwise. The proposal with the largest criteria score among the Top-K proposals is selected as the best-matched pair, providing a more reliable estimation of the object pose.

\vspace{-4mm}
\paragraph{Pose Estimation.}
With dense correspondences established across the best-matched views, we proceed to estimate the relative pose of the cameras. This pose estimation involves determining the rotation $\mathbf{R} \in \mathrm{SO}(3)$ and the translation vector $\mathbf{t} \in \mathbb{R}^3$ by matching descriptors, computing the essential matrix, and applying RANSAC to handle outliers and ensure reliable results~\cite{nister2004efficient}.
It is important to note that our method is capable of recovering the relative rotation accurately. However, the predicted translation is up-to-scale, similar to other relative pose estimator~\cite{gen6d,goodwin2022zero}. This limitation arises from the fact that recovering the absolute translation (or object scale) is an ill-posed problem when only considering two views, as it is susceptible to scale-translation ambiguity. To address this issue, we employ the PnP algorithm and utilize the bounding box of the prompted object in the uncropped support view to recovering the scale of translation.

\section{Experiments}
We initially demonstrate our approach for achieving zero-shot 6DoF object pose estimation on four different datasets using a two-view scenario. Subsequently, we validate the proposed open-world detector by assessing its segmentation and retrieval accuracy. 
Finally, in order to adapt POPE to multi-view pose estimation and evaluate the accuracy of multiple-view pose, we visualize the performance using additional input target frames and assess the pose on the task of novel view synthesis.

\begin{figure}
  \centering
  \includegraphics[width=0.9\linewidth]{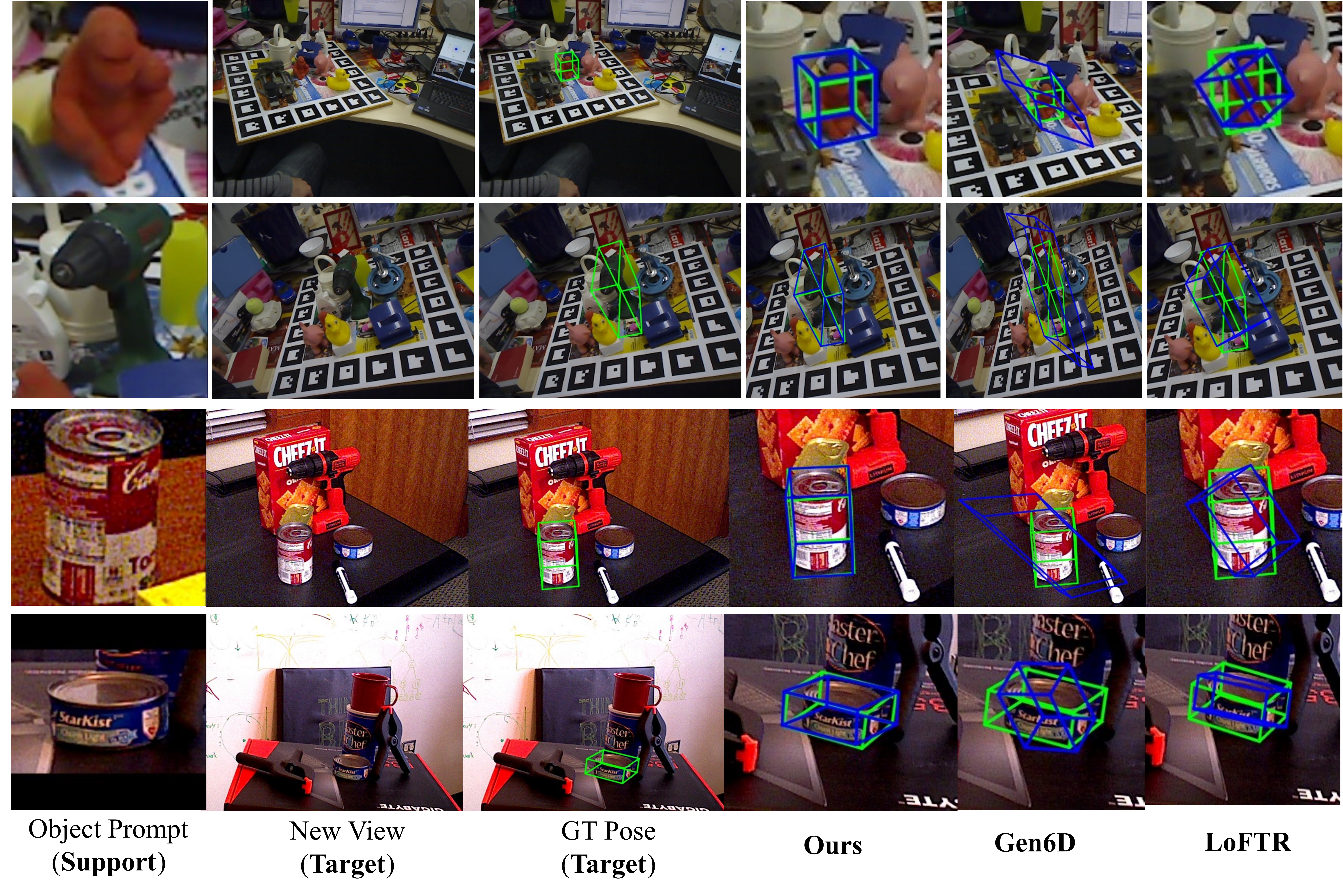}
  \vspace{-0.1in}
  \caption{\textbf{Qualitative results on the LINEMOD~\cite{hinterstoisser2013model} and YCB-Video~\cite{calli2015ycb} datasets.} Ground-truth poses are visualized with \textcolor{green}{green} boxes, while estimated poses are represented by \textcolor{blue}{blue} boxes. Gen6D performs poorly compared to correspondence-based methods, as it relies on a closed initial view for relative pose estimation. LoFTR tends to produce noisy matching results when using the object prompt and new view directly as input. POPE demonstrates robustness against complex and cluttered backgrounds by employing open-world object detection and finding correspondences through cropped and centralized images.}\label{fig:vis_linemod_ycb}
  \vspace{-2mm}
\end{figure}

\subsection{Evaluation Setup}
\paragraph{Datasets.}
We evaluate our method on four widely used 6DoF object pose estimation datasets, to test the zero-shot transferbility of \textbf{POPE} without any finetuning. \textbf{The LINEMOD Dataset~\cite{hinterstoisser2013model}} is a standard benchmark dataset for 6DoF object
pose estimation with the ground-truth CAD model. The LINEMOD dataset consists of images of thirteen low-textured objects under cluttered scenes and varying lighting conditions.
\textbf{The YCB-Video Dataset~\cite{calli2015ycb}} consists of 92 RGBD videos of 21 YCB objects, with medium clustered background and ground-truth CAD model for evaluation.
\textbf{The OnePose Dataset~\cite{sun2022onepose}} contains around 450 real-world video sequences of 150 objects with rich textures and simple background. Each frame is annotated with camera poses and 3D bounding box. 
\textbf{The OnePose++ Dataset~\cite{he2023onepose++}} supplements the original OnePose dataset with 40 household low-textured objects.
As the pose distribution in different datasets varies, we organize the manage the test set in a balanced distribution from $0^\circ$ to $30^\circ$.
Overall, the test set contains 5796 pairs on LIMEMOD, 2843 pairs on YCB-Video, 2751 pairs on OnePose, and 3166 pairs on OnePose++.
\vspace{-4mm}
\paragraph{Model Selection and Baselines.}
We compared our proposed POPE method with two other approaches: LoFTR~\cite{sun2021loftr}, an image-matching based method that directly performs correspondence matching for pose estimation, and Gen6D~\cite{gen6d}, which utilizes a correlation-based network to discover object boxes, find pose initialization, and refine the relative object pose.
We excluded the comparison with OnePose and OnePose++ as they are unable to generate point clouds from a single support view.
In POPE, we utilize pre-trained models for different tasks: the \textit{Segment Anything} model~\cite{kirillov2023segment} with a ViT-H architecture for object mask generation, the \textit{DINO-v2} model~\cite{oquab2023dinov2} pre-trained with ViT-S/14 for object proposal generation, and the \textit{LoFTR} model~\cite{sun2021loftr} pre-trained with indoor scenes for natural image-based image matching. We set $\sigma$ as 0.9 and the K as 3 in the experiments.
It is important to note that the evaluated promptable object pose estimation does not rely on labeled examples for fine-tuning, including the pose in the support view and object masks, for any objects in real-world environments.
\vspace{-4mm}
\paragraph{Evaluation.}
We report the median error for each pair of samples, along with the accuracy at 15$^\circ$ and 30$^\circ$, following the standard practice in relative object pose estimation~\cite{goodwin2022zero}. The accuracy metrics represent the percentage of predictions with errors below these thresholds. In the main draft, our evaluation primarily focuses on the two-view settings, while we provide additional results on downstream applications (multiple-view pose estimation, novel view synthesis).

\subsection{Comparisons}
\paragraph{Results on LINEMOD and YCB-video datasets.}
We present the overall average median error and pose accuracy under different thresholds in Table~\ref{tab:main}. Due to space limitations, we include the full table in the Sec~\ref{sec:num_each_instance} and demonstrate the median error for five instances in this section. It is evident from the results that the proposed \textbf{POPE} consistently outperforms other methods across all metrics, exhibiting a significant margin over each instance. The qualitative results, visualized in Figure~\ref{fig:vis_linemod_ycb}, highlight important observations. Gen6D~\cite{gen6d} heavily relies on accurate initialization for pose refinement and struggles in single-reference scenarios. LoFTR~\cite{sun2021loftr} fails to provide accurate matches when handling clustered scenes with object occlusions, resulting in inaccurate box predictions. It is important to note that the visualization of object boxes incorporates ground-truth translation to address scale ambiguity.

\begin{figure}\label{fig:vis_onepose_onepose++}
  \centering
  \includegraphics[width=0.9\linewidth]{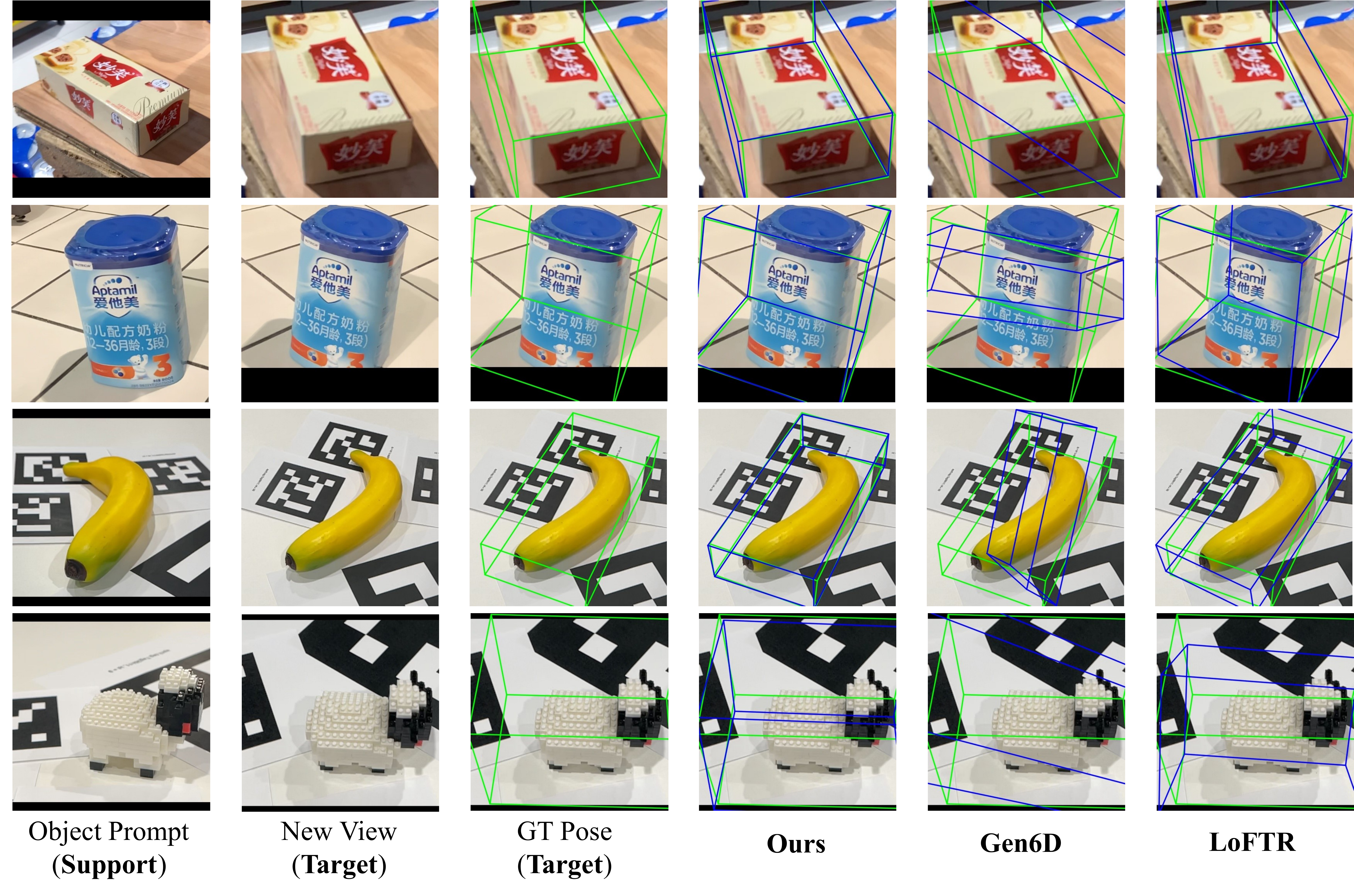}
  \vspace{-0.1in}
    \caption{\textbf{Qualitative results on the OnePose~\cite{sun2022onepose} and OnePose++~\cite{he2023onepose++} datasets.} Ground-truth poses are depicted with \textcolor{green}{green} boxes, while estimated poses are represented by \textcolor{blue}{blue} boxes. Gen6D performs poorly compared to correspondence-based methods due to the significant pose gap between the support and target views. LoFTR is susceptible to the presence of similar patterns between the object and background (last row). In contrast, our proposed POPE exhibits strong generalization ability on both textured and textureless single object datasets.}
    \vspace{2mm}
\end{figure}
\vspace{-4mm}
\paragraph{Results on OnePose and OnePose++ datasets.}
In addition to the dataset containing multiple objects in cluttered scenes, we also evaluate the proposed framework on recently introduced one-shot object pose estimation datasets. Unlike previous approaches that rely on pose or box annotations, we conduct zero-shot two-view pose estimation without such annotations. The results in Table~\ref{tab:main} demonstrate that \textbf{POPE} achieves a smaller median error in the relative object pose estimation task for both datasets. As the pose gap increases, LoFTR can improve its accuracy by utilizing the entire image for matching, incorporating more textural details from the background while still performing on par with our method.

\paragraph{Scaling from 2-view to Multi-view Promptable Pose Estimation (POPE)}
To address the requirement for sparse-view datasets in real-world scenarios, we have expanded our method from 2-view promptable pose estimation (POPE) to accommodate multi-view scenarios. Initially, we utilize the image matching results obtained from the 2-view POPE. We utilize the semi-dense correspondences from LOFTR~\cite{sun2021loftr} to reconstruct a semi-dense point cloud using COLMAP~\cite{schonberger2016structure}.

To introduce a new target viewpoint, we randomly select an image and perform object segmentation in a promptable manner. This enables us to retrieve the object's identity and exclude any negative effects caused by the clustered background. Subsequently, we conduct image matching between the prompt image and the newly added object image, register it, and extract correspondences between the new image and the semi-dense point cloud. The pose of the new object image is estimated by solving PnP. Finally, we update the sparse point cloud by minimizing reprojection errors and perform back-projection to obtain an optimized, accurate object point cloud, as well as updated object poses.

\begin{figure}[htbp]
  \centering
  \begin{subfigure}{0.49\textwidth}
    \includegraphics[width=\linewidth]{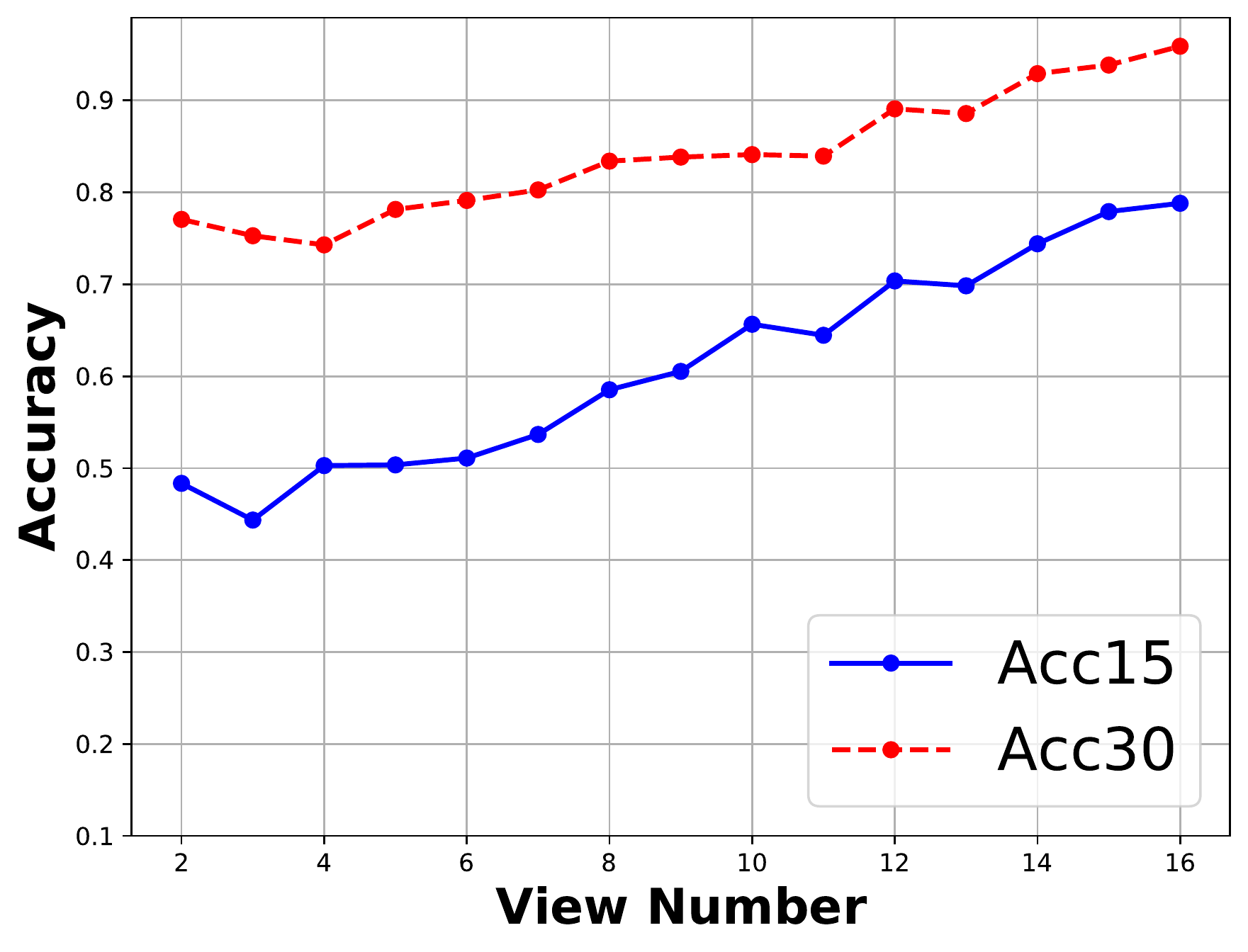}
    \caption{The accuracy under 15\textdegree (Acc15) and 30\textdegree (Acc30).}
    \label{fig:acc15_acc30}
  \end{subfigure}
  \hfill
  \begin{subfigure}{0.49\textwidth}
    \includegraphics[width=\linewidth]{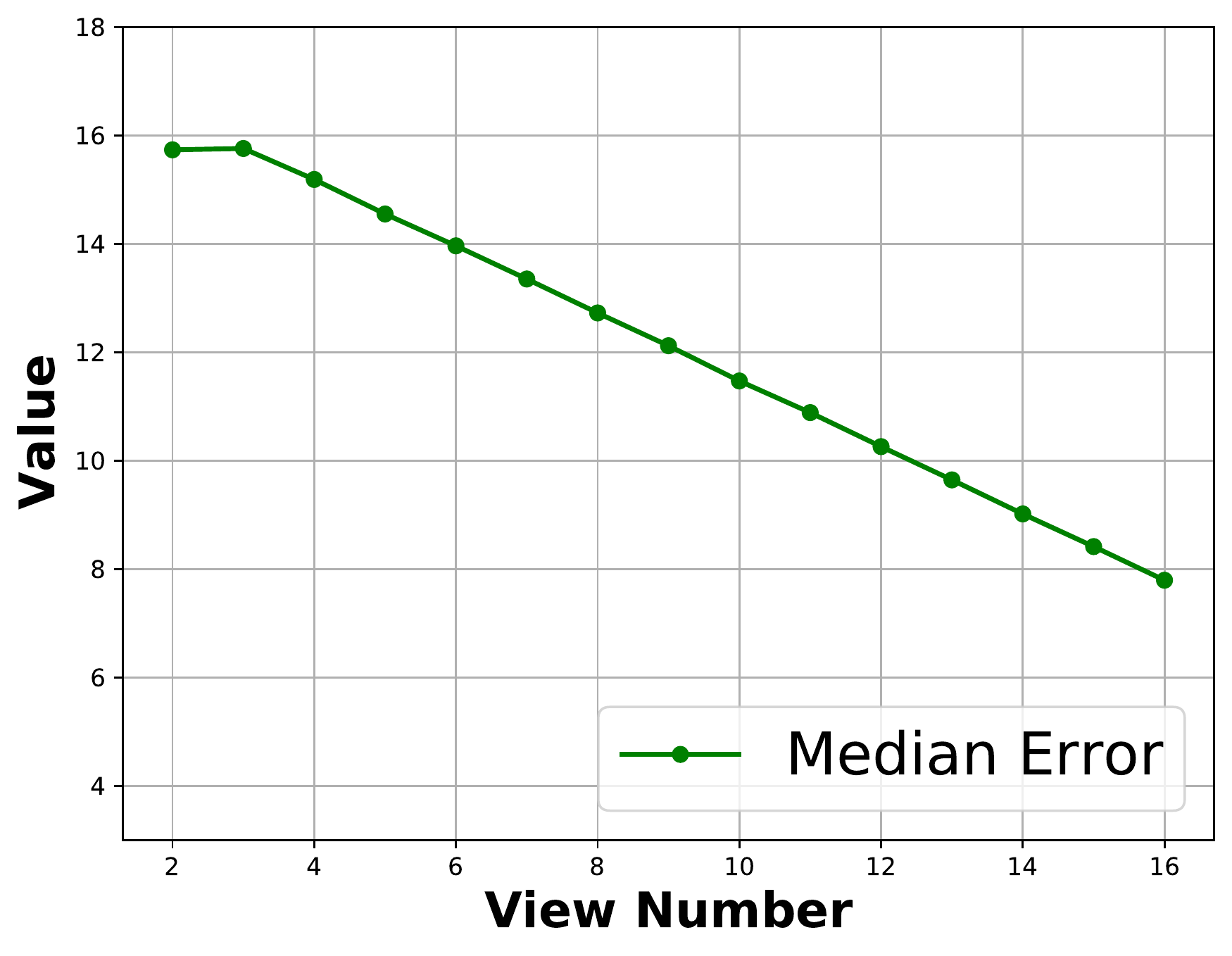}
    \caption{The Median Error with different view number.}
    \label{fig:acc15_acc30_median}
  \end{subfigure}
  \caption{We present plots illustrating the accuracy and median error as the number of views increases from 2 to 16.}
  \label{fig:acc_vs_view_num}
\end{figure}

To demonstrate the scalability of our method, we visualize the performance curve by randomly increasing the number of views. Figure~\ref{fig:acc_vs_view_num} illustrates that the overall accuracy significantly improves as more visual information is incorporated.

\begin{table}
\centering
\caption{We conduct experiments on zero-shot two-view object pose estimation on LINEMOD dataset, and report Median Error and Accuracy at 30º, 15º averaged across all 13 categories. We also report Accuracy at 30º broken down by class for an illustrative subset of categories.}
\label{tab:main}
\resizebox{0.99\linewidth}{!}{
\renewcommand{\arraystretch}{1.2}
\begin{tabular}{l|l|ccc|ccccc} 
\toprule
   \multirow{2}{*}{Dataset}  &  \multirow{2}{*}{Method} & \multicolumn{3}{c|}{All Categories} & \multicolumn{5}{c}{Per Category (Med. Err $\downarrow$)}  \\ 
\cline{3-10}
 &      & \multicolumn{1}{c}{Med. Err ($\downarrow$)} & \multicolumn{1}{c}{Acc30 ($\uparrow$)} & \multicolumn{1}{c|}{Acc15 ($\uparrow$)} & \multicolumn{1}{c}{Eggbox} & \multicolumn{1}{c}{Can} & \multicolumn{1}{c}{Iron} & \multicolumn{1}{c}{Hole.} & \multicolumn{1}{c}{Camera}  \\ 
\hline
\multirow{3}{*}{LINEMOD~\cite{hinterstoisser2013model}}   & Gen6D \cite{gen6d}    &  44.855             & 0.364                                       & 0.096                                        &   31.781                &   30.407                   &  30.094                      & 45.288                & 35.970                   \\
&   LoFTR \cite{sun2021loftr}        &33.036                                          &  0.562                              & 0.324                                   &16.887                  &17.585                   & 17.904                   &  31.782                       &   22.550                  \\
& Ours                                               & \textbf{15.731}                               & \textbf{0.770}                 & \textbf{0.483}                                 &  \textbf{10.530}           & \textbf{12.699}           &\textbf{13.157}          & \textbf{14.779}         & \textbf{15.102}            \\ 
\hline
      \multirow{4}{*}{YCB-Video~\cite{calli2015ycb}} &                &  Med. Err ($\downarrow$)        & Acc30 ($\uparrow$)        & Acc15 ($\uparrow$)         & 008                      & 003                        & 005                         & 006                             & 010                        \\ 
\cline{2-10}
 &Gen6D \cite{gen6d}                                             & 54.477                                     & 0.232                               &  0.077                             & 45.461                    &  80.992                   &  50.587                   &  66.999                              &    50.597                        \\
&LoFTR \cite{sun2021loftr}                                             &  19.5419                                   & 0.686                                 & 0.478                                  & 15.359             & 36.942              &   17.832               & 28.999                     &  17.475              \\
  & Ours                                             & \textbf{13.9411}                                   &  \textbf{0.801 }                             &  \textbf{0.544}                               & \textbf{7.787 }                 & \textbf{18.385}                     & \textbf{ 14.171}                  &\textbf{20.100}                            & \textbf{15.428}                 \\ 
\hline
        \multirow{4}{*}{OnePose++~\cite{he2023onepose++}} &              &  Med. Err ($\downarrow$)        & Acc30 ($\uparrow$)        & Acc15 ($\uparrow$)         &Mfmi.                      &  Oreo                       &  Taip.                        & Diyc.                             & Tee                       \\ 
\cline{2-10}
& Gen6D  \cite{gen6d}                                            & 35.428                                    & 0.411                                 & 0.158                                   &   16.963                  &   16.612                  & 17.787                     &19.132                      &  19.867                     \\
& LoFTR  \cite{sun2021loftr} &  9.012                                     & 0.891                                 &  0.703                                   &   4.077                 & 3.938                    & 4.147                     &  5.041                         &  5.312                   \\
&  Ours                                              & \textbf{6.273}                                     & \textbf{0.896}                                 & \textbf{0.728}                                 &    \textbf{1.765}               & \textbf{1.203}                    & \textbf{ 2.147}                     &   \textbf{2.769}                       &  \textbf{3.799}                  \\
\hline
\multirow{4}{*}{OnePose~\cite{sun2022onepose}}   &                      &  Med. Err ($\downarrow$)        & Acc30 ($\uparrow$)        & Acc15 ($\uparrow$)         &  Yell.                      &  Teab.                        & Oran.                         &  Gree.                             & Inst.                        \\ 
\cline{2-10}
 &   Gen6D \cite{gen6d}                                          & 17.785                                    & 0.893                                 & 0.389                                  &35.811         & 31.536        &   36.829                  &   30.609                  &    48.317                                      \\
& LoFTR \cite{sun2021loftr}                                             & 4.351                                      &   \textbf{0.963}                             &  \textbf{0.918}                                   & 9.773        &   6.488         &   9.439                    &  7.3482                   & 17.136                                    \\
 & Ours                                              &   \textbf{2.155}                                      &  0.962                                &  0.911                                 &  \textbf{5.470}                  &  \textbf{3.194}                   &   \textbf{8.044}                   &   \textbf{3.967}                        &  \textbf{16.492}                \\ 

\bottomrule
\end{tabular}}
\vspace{-4mm}
\end{table}

\vspace{-4mm}
\paragraph{Novel View Synthesis, an Application of POPE}
Our next objective is to validate the accuracy of our predicted pose estimation and demonstrate its practical applicability in downstream applications. To this end, we employ the estimated multi-view poses obtained from our POPE model, in combination with a pre-trained and generalizable Neural Radiance Field (GNT)~\cite{varma2022attention}.

\begin{figure}
  \centering
  \includegraphics[width=1\linewidth]{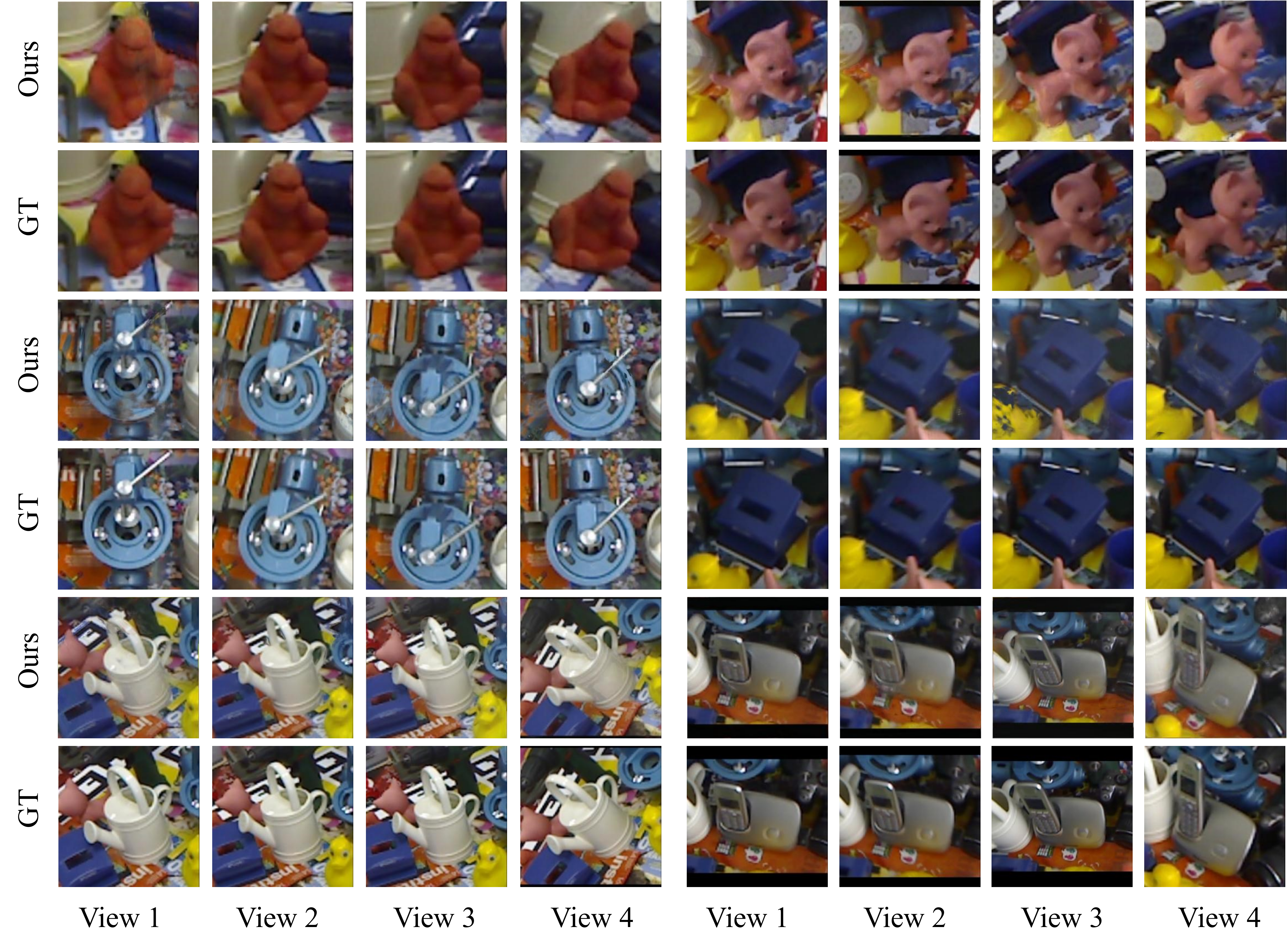}
  \caption{ \textbf{Application: Novel View Synthesis}. In the domain of novel view synthesis, we employ the poses obtained from our POPE model in conjunction with a pre-trained, generalizable Neural Radiance Field (GNT)~\cite{varma2022attention}. The GNT is configured with a source view number of 10, utilizing the ground truth poses for the purpose of warping. Subsequently, we generate 16 new viewpoints by rendering the scenes using the estimated poses derived from our POPE model.}\label{fig:nvs}
\end{figure}

Specifically, we configure the GNT with a source view number of 10 and utilize ground truth poses for source view warping. Subsequently, we leverage the estimated poses from our POPE model to generate new viewpoints based on the obtained POPE poses. Notably, our rendered results exhibit a remarkable resemblance to the ground-truth color image, as depicted in Figure~\ref{fig:nvs}, validating the precision of our estimated poses.

These findings provide compelling evidence supporting the accuracy and reliability of our pose estimation method, paving the way for its effective implementation in diverse downstream applications.

\vspace{-4mm}
\paragraph{Promptable Object Pose Estimation in Arbitrary Scene}

\begin{figure}[htbp]
  \centering
  \includegraphics[width=0.99\textwidth]{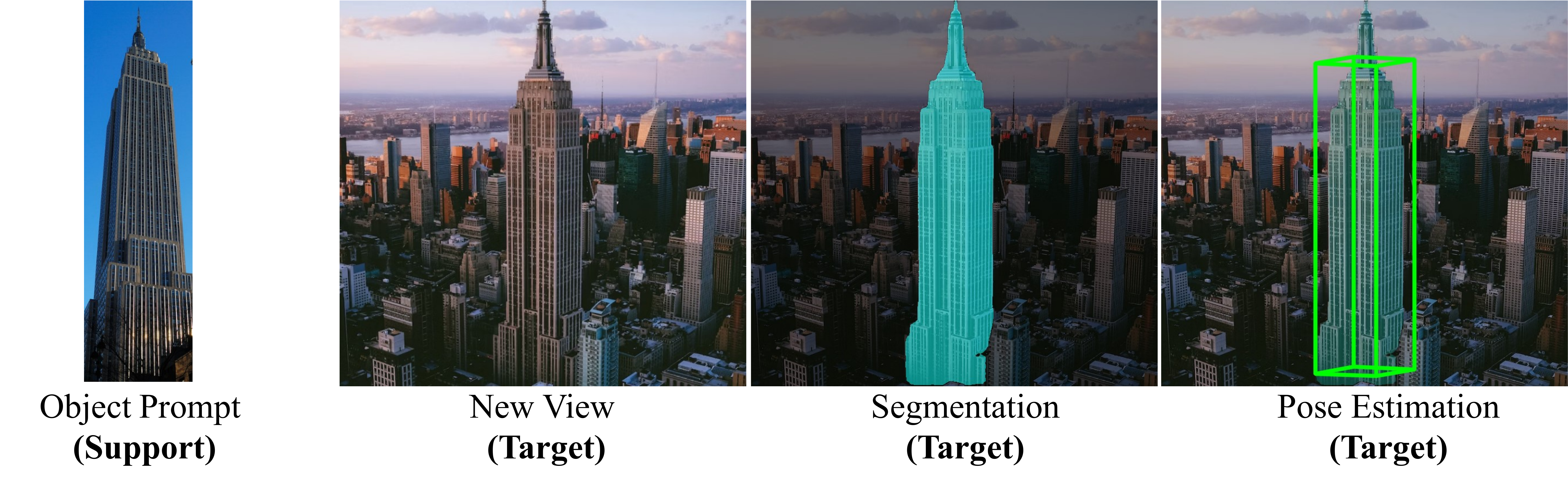}\hfill
  \includegraphics[width=0.99\textwidth]{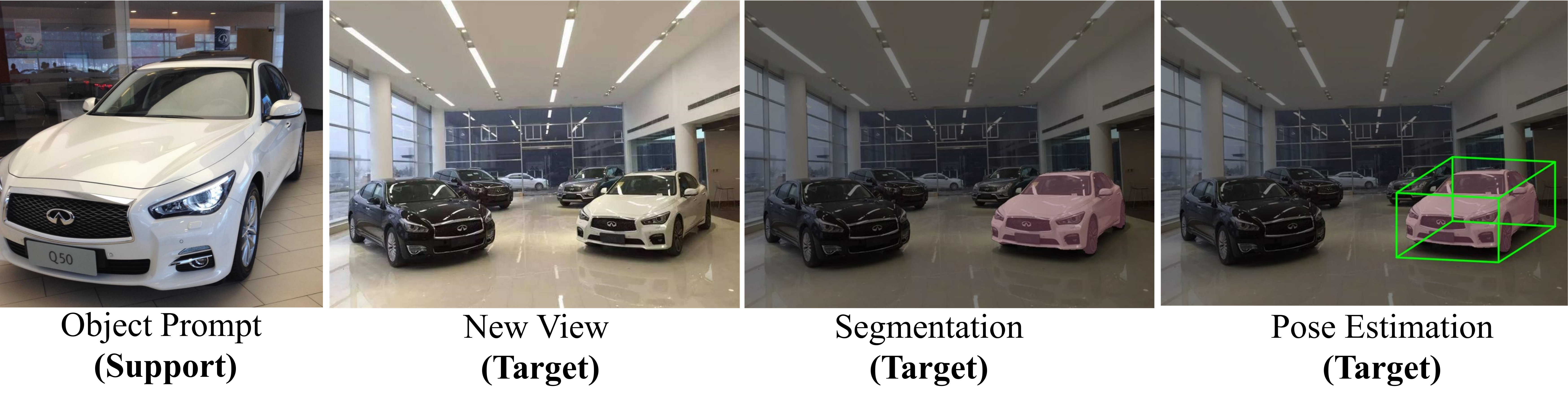}\\
  \caption{\textbf{Visual Examples of Promptable Object Pose Estimation for Outdoor Test Cases.} we present visual examples showcasing the retrieved object masks and the estimated relative poses in the context of promptable object pose estimation for outdoor test cases.
  }\label{fig:ood1}
\end{figure}

\begin{figure}[htbp]
  \centering
  \includegraphics[width=0.99\textwidth]{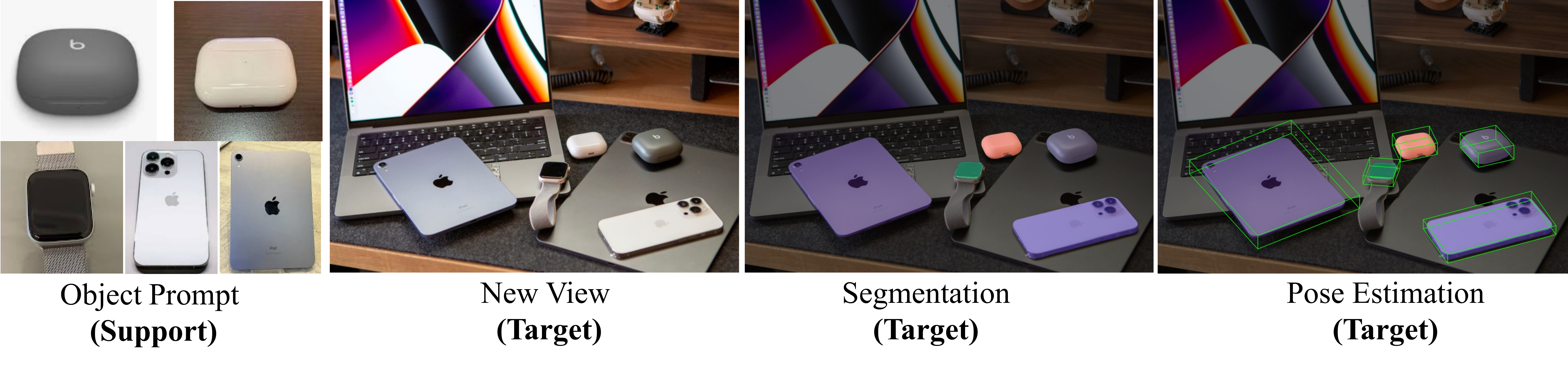}\hfill
  \includegraphics[width=0.99\textwidth]{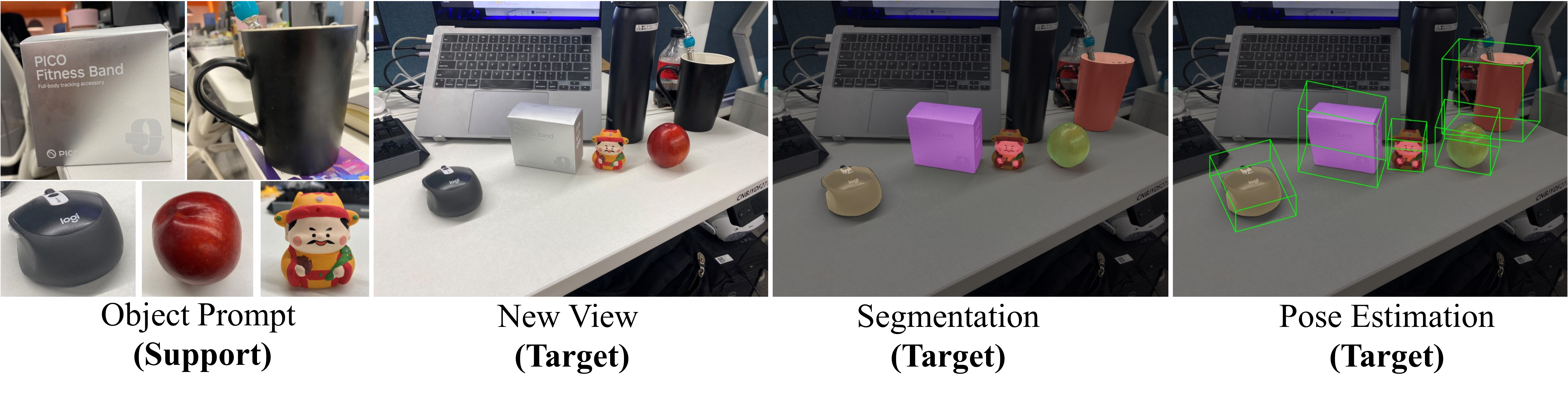}
  \caption{\textbf{Visual Examples of Promptable Object Pose Estimation for Indoor Test Cases.} we present visual examples showcasing the retrieved object masks and the estimated relative poses in the context of promptable object pose estimation for indoor test cases.
  }\label{fig:ood2}
\end{figure}

\begin{figure}[htbp]
  \centering
  \includegraphics[width=0.99\textwidth]{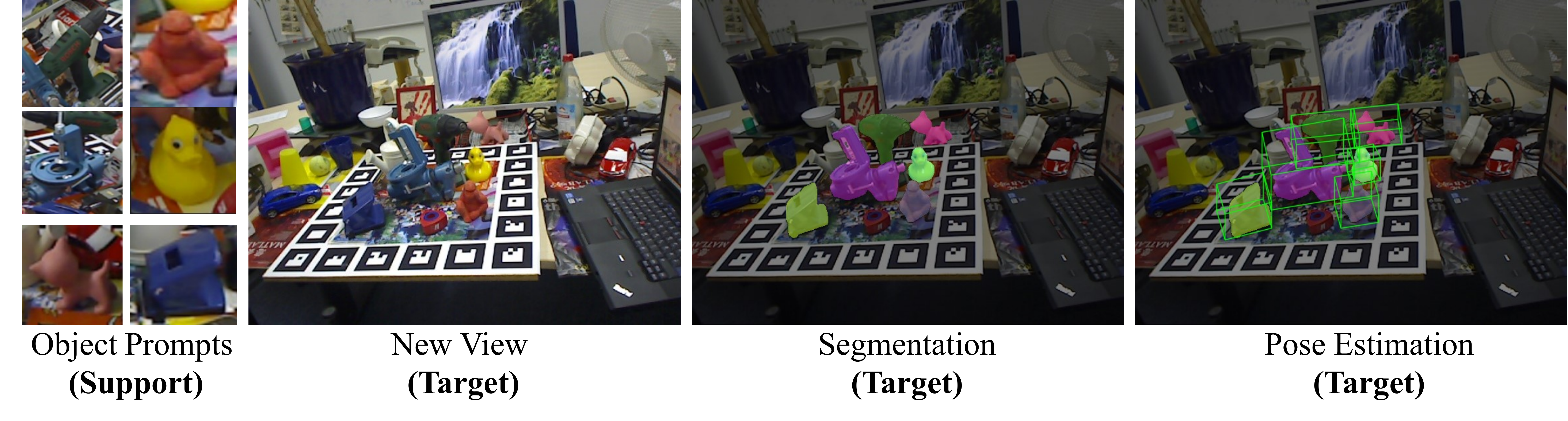}
  \caption{\textbf{Visual Examples of Promptable Object Pose Estimation for LINEMOD-Occ Test Cases.} we present visual examples showcasing the retrieved object masks and the estimated relative poses in the context of promptable object pose estimation on the occlusion subset of LINEMOD.
  }\label{fig:ood3}
\end{figure}

We provide supplementary visual examples in Figure~\ref{fig:ood1}, Figure~\ref{fig:ood2} and Figure~\ref{fig:ood3} to further illustrate the effectiveness of our promptable 6DoF object pose estimation method. This approach leverages a prompt image containing the object of interest, and our algorithm, \textbf{POPE}, demonstrates the ability to recognize objects of various categories through segmentation and retrieval processes, ultimately achieving accurate estimation of relative object poses.

\begin{figure}
  \centering
  \includegraphics[width=0.9\linewidth]{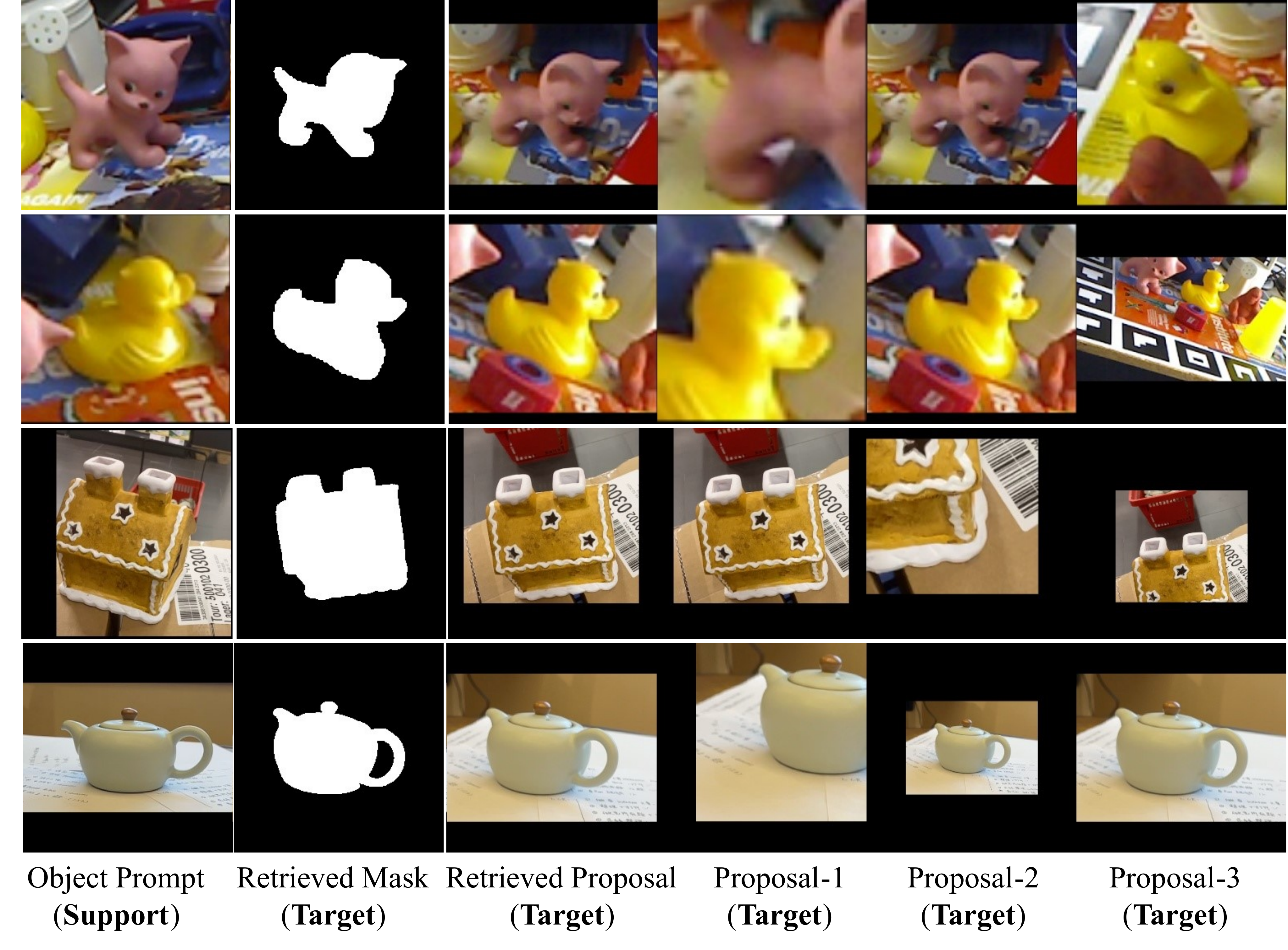}
  \vspace{-0.1in}
  \caption{\textbf{Ablation study}. Visualizations of retrieved object masks and proposals, selected from the Top-3 proposals using the global [CLS] token similarity.}\label{fig:ablation}
  \vspace{-5mm}
\end{figure}

\vspace{-4mm}
\paragraph{Necessity of Open-world Object Detector.}
Challenging scenes, such as cluttered or complex backgrounds, occlusions, or variations in lighting conditions, can pose significant challenges for traditional two-view object detection and pose estimation (see Tab~\ref{tab:main}). Whereas our proposed method utilize a open-world object detector, not limited to a specific group of classes, improves the zero-shot generalization by the retrieval-and-matching strategy. 
When retrieving using the global feature representation, which may mistakenly have large activations with non-related objects (Figure~\ref{fig:ablation}), results in the inaccurate 6DoF estimation in the later stage. 
The proposed hierarchical representation for object retrieval across viewpoints (Table~\ref{tab:abalation}), both improves the segmentation and retrieval accuracy, as well as benefits the subsequent pose estimation.

\begin{table}
\centering
\caption{\textbf{Ablation Studies.} We conducted an analysis of the segmentation, retrieval, and relative pose estimation tasks to validate the model design. The correlation-based detector in Gen6D~\cite{gen6d} often performs poorly in the clustered LINEMOD dataset when using only a single reference image (top row). The proposed framework, utilizing an \textit{Open-world Detector} that relies on global representation (second row), shows slightly lower performance compared to our full model, which incorporates hierarchical representation (last row). The results are averaged over a subset comprising 1/10 of the LINEMOD dataset.}
\label{tab:abalation}
\resizebox{0.9\linewidth}{!}{
\renewcommand{\arraystretch}{1.2}
\begin{tabular}{l|cc|c|ccc} 
\toprule
       \multirow{2}{*}{Method}               & \multicolumn{2}{c|}{Segmentation Acc.}                                         & Retrieval Acc. & \multicolumn{3}{c}{Pose Acc.}                                                   \\ 
\cline{2-7}
\multicolumn{1}{c|}{} & \multicolumn{1}{c}{mIoU ($\uparrow$)} & \multicolumn{1}{c|}{Accuracy($\uparrow$)} & mAP$\uparrow$             & \multicolumn{1}{c}{Med. Err$\downarrow$} & \multicolumn{1}{c}{Acc30$\uparrow$} & \multicolumn{1}{c}{Acc15$\uparrow$}  \\ 
\hline
Gen6D~\cite{gen6d}                 & 0.087                                & 0.102                                    &  0.067          &   44.644                   & 0.369                &  0.106                          \\
Ours(Global,Top-1)           &0.605                                & 0.815                                    & 0.817          &  14.912                   &   0.787               & 0.493                           \\
Ours(Hierarchical,Top-3)     &  \textbf{0.621}                              & \textbf{0.842}                               &   \textbf{0.844}        & \textbf{12.639}                 &  \textbf{0.810}                 &   \textbf{0.529}                     \\
\bottomrule
\end{tabular}}
\end{table}

\vspace{-4mm}
\paragraph{Quantitative Results on Each Instance}\label{sec:num_each_instance}
We provide a comprehensive analysis of the average median error and pose accuracy across various thresholds. Specifically, we present per-instance metrics for two-view 6DoF object pose estimation, focusing on datasets with clustered backgrounds, namely LINEMOD~\cite{hinterstoisser2013model} and YCB-Video~\cite{calli2015ycb}. The results are summarized in Table~\ref{tab:supp_linemod} and Table~\ref{tab:supp_ycb-video}, revealing a significant improvement in both per-instance accuracy and overall accuracy. This observation highlights the effectiveness of our promptable approach in mitigating the negative impact of background clutter and substantially enhancing the estimation accuracy.

\begin{table}[htbp]
\centering
\caption{We conduct experiments on zero-shot two-view object pose estimation on LINEMOD dataset, and report Median Error and Accuracy at 30º, 15º averaged across all 13 scenes.}
\label{tab:supp_linemod}
\resizebox{0.99\linewidth}{!}{
\begin{tabular}{l|l|cccccccccccccc} 
\toprule
\multirow{2}{*}{Metrics}                              & \multirow{2}{*}{Method} & \multicolumn{14}{c}{Per Instance}                                                                             \\ 
\cline{3-16}
                                                      &                         & ape    & benchvise & camera & can    & cat    & driller & duck   & eggbox & glue   & holepuncher & iron   & lamp   & phone  & Avg     \\ 
\hline
\multirow{3}{*}{Acc15 ($\uparrow$)}    & Gen6D                   & 0.016  & 0.125     & 0.112  & 0.120  & 0.028  & 0.157   & 0.029  & 0.114  & 0.023  & 0.0670      & 0.108  & 0.241  & 0.105  & 0.096   \\
                                                      & LoFTR                   & 0.091  & 0.423     & 0.338  & 0.429  & 0.172  & 0.445   & 0.190  & 0.433  & 0.119  & 0.253       & 0.411  & 0.582  & 0.322  & 0.324   \\
                                                      & Ours                    & 0.439  & 0.450     & 0.493  & 0.531  & 0.444  & 0.47916 & 0.456  & 0.607  & 0.380  & 0.502       & 0.585  & 0.467  & 0.445  & 0.483   \\ 
\hline
\multirow{3}{*}{Acc30 ($\uparrow$)}      & Gen6D                   & 0.133  & 0.445     & 0.400  & 0.485  & 0.232  & 0.482   & 0.203  & 0.437  & 0.147  & 0.279       & 0.496  & 0.609  & 0.380  & 0.364   \\
                                                      & LoFTR                   & 0.291  & 0.663     & 0.608  & 0.7125 & 0.388  & 0.687   & 0.370  & 0.722  & 0.248  & 0.480       & 0.738  & 0.855  & 0.542  & 0.562   \\
                                                      & Ours                    & 0.789  & 0.710     & 0.764  & 0.826  & 0.732  & 0.765   & 0.758  & 0.840  & 0.686  & 0.809       & 0.857  & 0.733  & 0.743  & 0.770   \\ 
\hline
\multirow{3}{*} {Med. Err ($\downarrow$)}     & Gen6D                   & 79.705 & 32.504    & 35.970 & 30.407 & 54.468 & 30.665  & 57.292 & 31.781 & 88.044 & 45.288      & 30.094 & 25.551 & 39.392 & 44.855  \\
                                                      & LoFTR                   & 70.094 & 19.227    & 22.550 & 17.585 & 43.069 & 18.356  & 44.083 & 16.887 & 90.000 & 31.782      & 17.904 & 11.871 & 26.063 & 33.036  \\
                                                      & Ours                    & 16.716 & 17.762    & 15.102 & 12.699 & 17.921 & 15.926  & 17.641 & 10.530 & 19.144 & 14.779      & 13.157 & 16.203 & 16.929 & 15.731  \\
\bottomrule
\end{tabular}}
\end{table}

\begin{table}[htbp]
\centering
\caption{We conduct experiments on zero-shot two-view object pose estimation on YCB-Video dataset, and report Median Error and Accuracy at 30º, 15º averaged across all 10 scenes.}
\label{tab:supp_ycb-video}
\resizebox{0.99\linewidth}{!}{
\begin{tabular}{l|l|ccccccccccc} 
\toprule
\multirow{2}{*}{Metrics}                                  & \multirow{2}{*}{Method} & \multicolumn{11}{c}{Per Instance}                                                                 \\ 
\cline{3-13}
                                                          &                         & 001    & 002    & 003    & 004    & 005    & 006    & 007    & 008    & 009    & 010    & Avg     \\ 
\hline
\multirow{3}{*}{Acc15 ($\uparrow$)}      & Gen6D                   & 0.046  & 0.063  & 0.028  & 0.017  & 0.084  & 0.027  & 0.250  & 0.102  & 0.073  & 0.085  & 0.077   \\
                                                          & LoFTR                   & 0.483  & 0.539  & 0.297  & 0.245  & 0.457  & 0.298  & 1.000  & 0.4953 & 0.508  & 0.457  & 0.478   \\
                                                          & Ours                    & 0.441  & 0.547  & 0.401  & 0.457  & 0.521  & 0.381  & 0.937  & 0.738  & 0.524  & 0.492  & 0.544   \\ 
\hline
\multirow{3}{*}{Acc30 ($\uparrow$)}      & Gen6D                   & 0.204  & 0.190  & 0.140  & 0.108  & 0.253  & 0.138  & 0.562  & 0.308  & 0.221  & 0.192  & 0.232   \\
                                                          & LoFTR                   & 0.637  & 0.817  & 0.481  & 0.485  & 0.739  & 0.506  & 1.000  & 0.785  & 0.6885 & 0.721  & 0.686   \\
                                                          & Ours                    & 0.655  & 0.857  & 0.755  & 0.748  & 0.816  & 0.680  & 1.000  & 0.953  & 0.778  & 0.771  & 0.801   \\ 
\hline
\multirow{3}{*}{Med. Err ($\downarrow$)} & Gen6D                   & 53.87  & 49.995 & 80.992 & 64.819 & 50.587 & 66.999 & 27.633 & 45.461 & 53.817 & 50.597 & 54.477  \\
                                                          & LoFTR                   & 17.198 & 13.484 & 36.942 & 31.474 & 17.832 & 28.999 & 2.038  & 15.359 & 14.613 & 17.475 & 19.541  \\
                                                          & Ours                    & 18.582 & 12.133 & 18.385 & 17.257 & 14.171 & 20.100 & 1.408  & 7.7875 & 14.156 & 15.428 & 13.941  \\
\bottomrule
\end{tabular}}
\end{table}

\begin{table}[htbp]
\centering
\caption{We conduct experiments on zero-shot two-view object pose estimation on OnePose dataset, and report Median Error and Accuracy at 30º, 15º averaged across all 10 objects.} 
\label{tab:supp_onepose}
\resizebox{0.99\linewidth}{!}{
\begin{tabular}{l|l|ccccccccccc} 
\toprule
\multirow{2}{*}{Metrics}                                  & \multirow{2}{*}{Method} & \multicolumn{11}{c}{Per Instance}                                                                                               \\ 
\cline{3-13}
                                                          &                         & aptamil & jzhg   & minipuff & hlyormosiapie & brownhouse & oreo   & mfmilkcake & diycookies & taipingcookies & tee    & Avg     \\ 
\hline
\multirow{3}{*}{Acc15 ($\uparrow$)}      & Gen6D                   & 0.350   & 0.445  & 0.387    & 0.397         & 0.424      & 0.421  & 0.417      & 0.357      & 0.394          & 0.299  & 0.389   \\
                                                          & LoFTR                   & 0.872   & 0.931  & 0.964    & 0.897         & 0.984      & 0.957  & 0.947      & 0.822      & 0.975          & 0.834  & 0.918   \\
                                                          & Ours                    & 0.871   & 0.959  & 0.925    & 0.886         & 0.968      & 0.975  & 0.920      & 0.8        & 0.963          & 0.849  & 0.911   \\ 
\hline
\multirow{3}{*}{Acc30 ($\uparrow$)}      & Gen6D                   & 0.845   & 0.914  & 0.925    & 0.901         & 0.944      & 0.914  & 0.923      & 0.796      & 0.938          & 0.831  & 0.893   \\
                                                          & LoFTR                   & 0.945   & 0.982  & 0.982    & 0.978         & 0.992      & 0.992  & 0.969      & 0.878      & 0.993          & 0.918  & 0.963   \\
                                                          & Ours                    & 0.949   & 0.979  & 0.973    & 0.974         & 0.976      & 0.985  & 0.967      & 0.895      & 0.993          & 0.930  & 0.962   \\ 
\hline
\multirow{3}{*}{Med. Err ($\downarrow$)} & Gen6D                   & 19.542  & 16.356 & 17.348   & 17.500        & 16.747     & 16.612 & 16.963     & 19.132     & 17.787         & 19.867 & 17.785  \\
                                                          & LoFTR                   & 5.407   & 4.182  & 3.978    & 3.869         & 3.555      & 3.938  & 4.077      & 5.041      & 4.147          & 5.312  & 4.351   \\
                                                          & Ours                    & 2.997   & 1.460  & 1.786    & 2.155         & 1.470      & 1.2033 & 1.765      & 2.769      & 2.147          & 3.799  & 2.155   \\
\bottomrule
\end{tabular}}
\end{table}

\begin{table}[htbp]
\centering
\caption{We conduct experiments on zero-shot two-view object pose estimation on OnePose++ dataset, and report Median Error and Accuracy at 30º, 15º averaged across all 9 objects.}
\label{tab:supp_onepose++}
\resizebox{0.99\linewidth}{!}{
\begin{tabular}{l|l|cccccccccc} 
\toprule
\multirow{2}{*}{Dataset}                                  & \multirow{2}{*}{Method} & \multicolumn{10}{c}{Per Instance}                                                                             \\ 
\cline{3-12}
                                                          &                         & toyrobot & yellowduck & sheep  & fakebanana & teabox & orange & greenteapot & lecreusetcup & insta  & Avg     \\ 
\hline
\multirow{3}{*}{Acc15 ($\uparrow$)}      & Gen6D                   & 0.171    & 0.123      & 0.197  & 0.156      & 0.204  & 0.135  & 0.185       & 0.185        & 0.067  & 0.158   \\
                                                          & LoFTR                   & 0.794    & 0.676      & 0.772  & 0.68       & 0.782  & 0.685  & 0.783       & 0.708        & 0.443  & 0.703   \\
                                                          & Ours                    & 0.753    & 0.768      & 0.781  & 0.683      & 0.844  & 0.7    & 0.860       & 0.708        & 0.460  & 0.728   \\ 
\hline
\multirow{3}{*}{Acc30 ($\uparrow$)}      & Gen6D                   & 0.451    & 0.361      & 0.472  & 0.423      & 0.478  & 0.388  & 0.479       & 0.413        & 0.232  & 0.411   \\
                                                          & LoFTR                   & 0.912    & 0.901      & 0.922  & 0.893      & 0.903  & 0.855  & 0.969       & 0.928        & 0.738  & 0.891   \\
                                                          & Ours                    & 0.882    & 0.936      & 0.901  & 0.88       & 0.919  & 0.905  & 0.953       & 0.907        & 0.781  & 0.896   \\ 
\hline
\multirow{3}{*}{Med. Err ($\downarrow$)} & Gen6D                   & 32.998   & 35.811     & 31.366 & 36.202     & 31.536 & 36.829 & 30.609      & 35.185       & 48.317 & 35.428  \\
                                                          & LoFTR                   & 6.368    & 9.773      & 7.336  & 8.751      & 6.488  & 9.439  & 7.348       & 8.472        & 17.136 & 9.012   \\
                                                          & Ours                    & 3.792    & 5.470      & 4.435  & 4.990      & 3.194  & 8.044  & 3.967       & 6.072        & 16.492 & 6.273   \\
\bottomrule
\end{tabular}}
\end{table}
Furthermore, we present per-instance metrics for two-view 6DoF object pose estimation on datasets containing single object with rich textures~\cite{sun2022onepose} and poor textures~\cite{he2023onepose++} of each scebe. As depicted in Table~\ref{tab:supp_onepose} and Table~\ref{tab:supp_onepose++}, our method outperforms other two-view-based methods in terms of pose accuracy, with assistance of foreground object segmentation and retrieval.

\section{Conclusion}
In this paper, we present Promptable Object Pose Estimator (POPE), a zero-shot solution for estimating the 6DoF object pose in any scene with only one support image. Our solution highlights the potential of leveraging 2D pre-trained foundation models to lift the typical object pose estimation to generalize in a more practical paradigm. It features a modular design that decomposes the promptable object pose estimation into several steps. We demonstrate the scalability of our proposed solution to use single support image as prompt under extreme clustered scenes, the extension to multiple viewpoints, and the validation on novel view synthesis.
Several potential directions for future work exist, including distilling large-scale foundation models into smaller ones for enabling real-time inference, and incorporating single-view depth information from a monocular depth estimator to enhance zero-shot accuracy. We envision that our solution will enable users to generate photorealistic 3D assets for augmented or virtual reality applications using only a few images, even as sparse as two.

\bibliographystyle{unsrt}
\bibliography{reference}

\end{document}